# Deep Learning for Medical Image Processing: Overview, Challenges and Future

Muhammad Imran Razzak, Saeeda Naz and Ahmad Zaib

Abstract :

   Healthcare sector is totally different from other industry. It is on high priority sector and people expect highest level of care and services regardless of cost. It did not achieve social expectation even though it consume huge percentage of budget. Mostly the interpretations of medical data is being done by medical expert. In terms of image interpretation by human expert, it is quite limited due to its subjectivity, complexity of the image, extensive variations exist across different interpreters, and fatigue. After the success of deep learning in other real world application, it is also providing exciting solutions with good accuracy for medical imaging and is seen as a key method for future applications in health secotr. In this chapter, we discussed state of the art deep learning architecture and its optimization used for medical image segmentation and classification. In the last section, we have discussed the challenges deep learning based methods for medical imaging and open research issue.

## 1 Introduction

Gone are the days, when health-care data was small. Due to the tremendous advancement in image acquisition devices, the data is quite large (moving to big data), that makes it challenging and interesting for image analysis. This rapid growth in medical images and modalities requires extensive and

_____________________

*M.I. Razzak*
*CPHHI, KSAU-HS, Riyadh e-mail: mirpake@gmail.com*

*Saeeda Naz*
*GGPGC1, Abbottabad e-mail: saeedanaz291@gmail.com*

*Ahmad Zaib*
*Women Medical College, Abbottabad e-mail: ahmad.zaib@gmail.com*

tedious efforts by medical expert that is subjective, prone to human error and may have large variations across different expert. Alternative solution is using machine learning techniques to automate diagnosis process however, traditional machine learning methods are not sufficient to deal with com-plex problem. Happy marriage of high performance computing with machine learning promise the capacity to deal big medical image data for accurate and efficient diagnosis. Deep learning will not only help to not only help to select and extract features but also construct new ones, furthermore, it does not only diagnose the disease but also measure predictive target and provides actionable prediction models to help physician efficiently.

Machine Learning (ML) and Artificial Intelligence (AI) have progressed rapidly in recent years. Techniques of ML and AI have played important role in medical field like medical image processing, computer-aided diagnosis, image interpretation, image fusion, image registration, image segmentation, image-guided therapy, image retrieval and analysis Techniques of ML extract information from the images and represents information effectively and effi-ciently. The ML and AI facilitate and assist doctors that they can diagnose and predict accurate and faster the risk of diseases and prevent them in time. These techniques enhance the abilities of doctors and researchers to under-stand that how to analyze the generic variations which will lead to disease. These techniques composed of conventional algorithms without learning like Support Vector Machine (SVM), Neural Network (NN), KNN etc. and deep learning algorithms such as Convolutional Neural Network (CNN), Recur-rent neural Network (RNN), Long Short term Memory (LSTM), Extreme Learning Model (ELM), Generative Adversarial Networks (GANs) etc. For-mer algorithms are limited in processing the natural images in their raw form, time consuming, based on expert knowledge and requires a lot time for tuning the features. The later algorithms are fed with raw data, automatic features learner and fast. These algorithms try to learn multiple levels of abstraction, representation and information automatically from large set of images that exhibit the desired behavior of data. Although automated detec-tion of diseases based on conventional methods in medical imaging has been shown significant accuracies around for decades, but new advances in machine learning techniques have ignited a boom in the deep learning. Deep learning based algorithms showed promising performance as well speed in different do-mains like speech recognition, text recognition, lips reading, computer-aided diagnosis, face recognition, drug discovery.

The motivation of this chapter is to provide the comprehensive review of deep learning based algorithms in medical image analysis problems in terms of current work and future direction. This chapter provides the fundamental knowledge and the state of the art approaches about deep learning in the domain of medical image processing and analysis.

## 2 Why Deep Learning Over Machine Learning

Accurate diagnoses of disease depends upon image acquisition and image interpretation. Image acquisition devices has improved substantially over the recent few years i.e. currently we are getting radiological images ((X-Ray, CT and MRI scans etc.) with much higher resolution. However, we just started to get benefits for automated image interpretation. One of the best machine learning application is computer vision, though traditional machine learning algorithms for image interpretation rely heavily on expert crafted features i.e. lungs tumor detection requires structure features to be extracted. Due to the extensive variation from patient to patient data, traditional learning methods are not reliable. Machine learning has evolved over the last few years by its ability to shift through complex and big data.

Now deep learning has got great interest in each and every field and especially in medical image analysis and it is expected that it will hold $300 million medical imaging market by 2021. Thus, by 2021, it alone will get more more investment for medical imaging than the entire analysis industry spent in 2016. It is the most effective and supervised machine learning approach. This approach use models of deep neural network which is variation of Neural Network but with large approximation to human brain using advance mechanism as compare to simple neural network. The term deep learning implies the use of a deep neural network model. The basic computational unit in a neural network is the neuron, a concept inspired by the study of the human brain, which takes multiple signals as inputs, combines them linearly using weights, and then passes the combined signals through nonlinear operations to generate output signals.

## 2.1 Neural Network and Deep Learning Architecture

Artificial neural networks structurally and conceptually inspired by human biological nervous system . Preceptron is one of the earliest neural network that was based on human brain system. It consist of input layer that is directly connect to output layer and was good to classify linearly separable patterns. To solve more complex pattern, neural network was introduced that has a layered architecture i.e., input layer, output layer and one or more hidden layers. Neural network consist of interconnected neurons that takes input and perform some processing on the input data, and finally forward the current layer output to the coming layer. The general architecture of neural network is shown in figure 2. Each neuron in the network sums up the input data and apply the activation function to the summed data and finally pro-vides the output that might be propagated to the next layer. Thus adding more hidden layer allows to deal with complex as hidden layer capture non-linear relationship. These neural networks are knows as Deep Neural network. Deep learning provides new cost effective to train DNN were slow in learning the weights. Extra layers in DNN enable composition of features from lower layers to the upper layer by giving the potential of modeling complex data.

Deep learning is the growing trend to develop automated applications and has been termed in 10 breakthrough technologies of 2013. Today, several deep learning based computer vision applications are performing even better than human i.e. identifying indicators for cancer in blood and tumors in MRI scans. It is improvement of artificial neural network that consist of more hidden layer that permits higher level of abstraction and improved image analysis. It becomes extensively applied method due to its recent unparal-lelled result for several applications i.e. object detection, speech recognition, face recognition and medical imaging.

A deep neural network hierarchically stacks multiple layers of neurons, forming a hierarchical feature representation. The number of layers now extends to over 1,000! With such a gigantic modeling capacity, a deep network can essentially memorize all possible mappings after successful training with a sufficiently large knowledge database and make intelligent predictions e.g. interpolations and/or extrapolations for unseen cases. Thus, deep learning is generating a major impact in computer vision and medical imaging. In fact, similar impact is happening in domains like text, voice, etc. Various types of deep learning algorithms are in use in research like convolutional neural networks (CNN), deep neural network (DNN), deep belief network (DBN), deep autoencodre (dA), deep Boltzmann machine (DBM), deep conventional extreme machine learning (DC-ELM) recurrent neural network (RNN) and its variant like BLSTM and MDLATM etc (as illustrated with their pros and cons in table 1. The CNN model is getting a lot interest in digital imaging processing and vision. There different types of architectures of CNN such as, Alexnet[1] (as shown in Figure 10, Lenet[2], faster R-CNN[3], googleNEt[4], ResNEt[5], VGGNet[6], ZFnet etc.

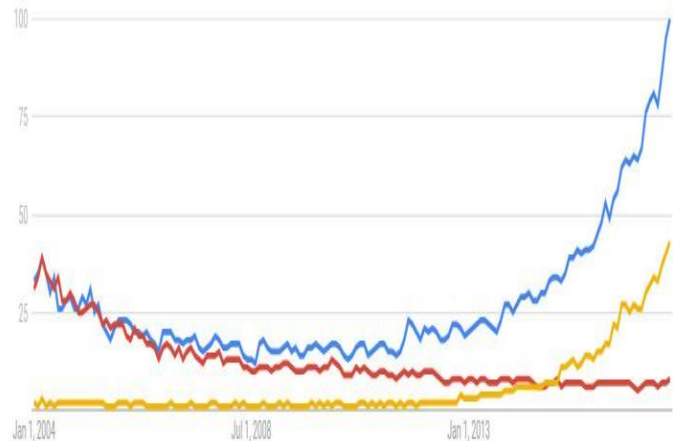

**Fig. 1: Trends: Deep Learning vs Machine Learning vs Pattern Recognition**

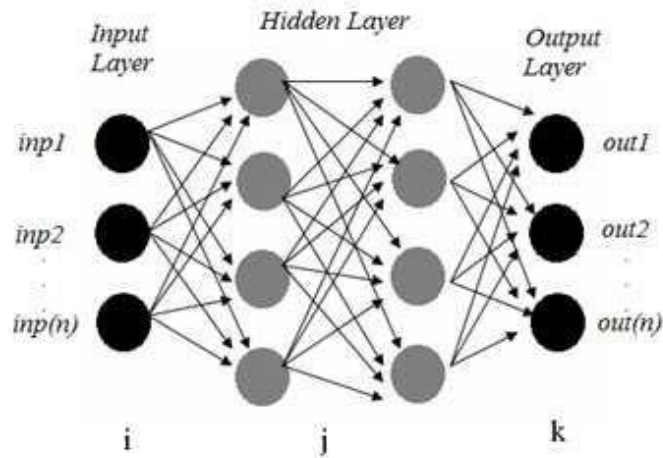

**Fig. 2: Neural network architecture**

[1] *https://github.com/BVLC/caffe/tree/master/models/bvlc*alexnet

[2] *http://deeplearning.net/tutorial/lenet.html*

[3] *https://github.com/ShaoqingRen/faster*r cnn

[4] *https://github.com/BVLC/caffe/tree/master/models/bvlc*g ooglenet

[5] *https://github.com/gcr/torch-residual-networks*

[6] *http://www.robots.ox.ac.uk/ vgg/research/very*deep/

Table 1: Comparison of different architecture of Deep learning models

| Type of Network | Detail of Network | PROS | CONS |
|---|---|---|---|
| Deep Neural Network (DNN) | There are more than two layers. which allow complex non-linear relationship.It is used for classification as well for regression. The architecture is shown. in Figure 5. | It is widely used with great accuracy | the training process is not trivial because the error is propagated back to the previous one layers and they become very small. The learning process of the model is also too much slow. |
| Convolutional Neural Network (CNN) | This network is very good for 2 dimensional data. It consists of convolutional filters which transform 2D into 3D. The architecture has shown in Figure 3 | very good performance, learning of model is fast | It needs a lot of labeled data for classification. |
| Recurrent Neural Network (RNN) | It has the capability of learning of sequences. the wights are sharing across all steps and neurons | learn sequential events, can model time dependencies, there are many variation like LSTM, BLSTM, MDLSTM, HLSTM. These provide state of the art accuracies in speech recognition, character recognition and many other NLP related tasks | there many issue due to gradient vanishing and need of big datasets |
| Deep Conventional Extreme Learning Machine (DC-ELM) | For sampling of local connections, this network uses Gaussian probability function as shown in Figure ?? | | |
| | | | Continued on next page |

Table 1 – continued from previous page

| Type of Network | Detail of Network | PROS | CONS |
| --- | --- | --- | --- |
| Deep Boltzmann Machine (DBM) | This model is based on family of Boltzmann and it consists of unidirectional connections between all hidden layers as shown in Figure ?? | the top-down feedback incorporates with ambiguous data for more robust inference | optimization of parameters is not possible for big dataset. |
| Deep Belief Network (DBN) | This model has unidirectional connection at two layers on the top of layers. It is used in both supervised and supervised learning in machine learning. The hidden layers of each sub-network serves as visible layer for the next layer. The architecture is shown in Figure ?? | The greedy strategy used in each layer and the inference tractable maximize directly the likelihood | The initialization make the training process computationally expensive |
| Deep Auto-encder (dA) | It is used in unsupervised learning and it is designed mainly for extraction and reduction of dimensionality of features. The number of input is equal to number of output. It is shown in Figure 7 | It does not need labelled data. there are different variations like Sparse Auto-encoder, De-nosing Auto-encoder, Conventional Auto-Enc for more robustness. | IT needs pre-training step. Its training may suffer from vanishing. |

## 3 Deep Learning: Not-so-near Future in Medical Imaging

Deep learning technology applied to medical imaging may become the most disruptive technology radiology has seen since the advent of digital imag-ing. Most researchers believe that within next 15 years, deep learning based applications will take over human and not only most of the diagnosis will be performed by intelligent machines but will also help to predict disease, prescribe medicine and guide in treatment. Which field in medical has rev-olutionised Deep learning first? ophthalmology,pathology, cancer detection, radiology or prediction and personalized medicine. Ophthalmology will be the first field to be revolutionized in health care, however, pathology, cancer di-agnosis have received more attention and currently we have application with decent accuracy. Google DeepMind Health is working with National Health Service, UK signed five year agreement to process the medical data of up to 1m patients across the trusts five hospitals. Even its early days of this project, Deepmind already has high hopes for the proposal.

Researchers and vendors in medical sector are moving this field forward have a bold recommendation i.e. IBM Watson recently boosted itself through billion-dollar entry into the imaging arena by the acquisition Merge (imag-ing and Google DeepMind Health is another big investment. Even though, huge investment and interest, deep learning future in medical imaging is not that near as compare to other imaging applications due to the complexities involved in this field. The notion of applying deep learning based algorithms to medical imaging data is a fascinating and growing research area however, there are several barriers that slow down its progress. These challenges are unavailability of dataset, privacy and legal issues, dedicated medical experts, non standard data machine learning algorithms etc.

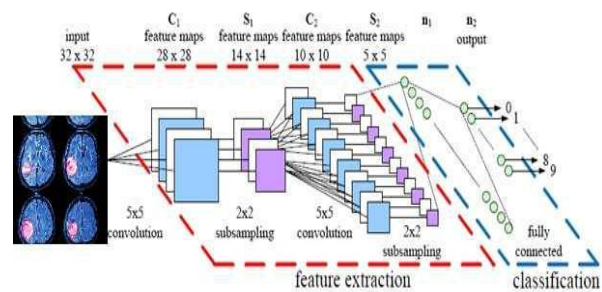

**Fig. 3: Convolutional Neural Network**

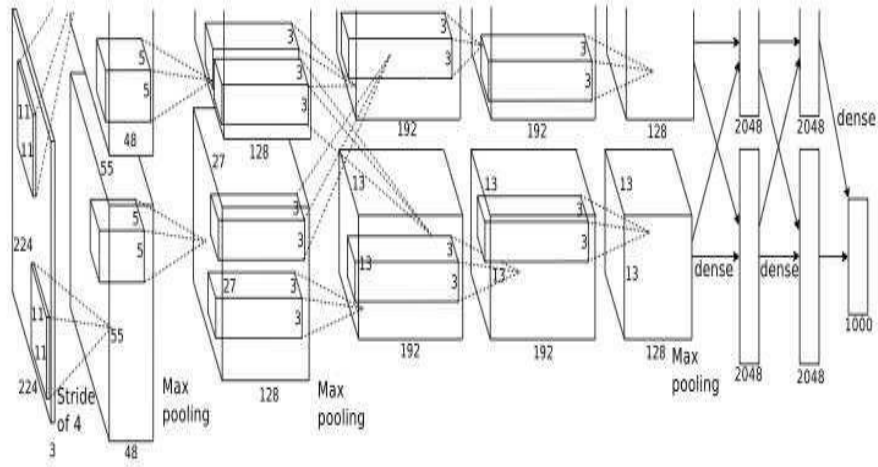

*Fig. 4: type of architecture of CNN:AlexNet*

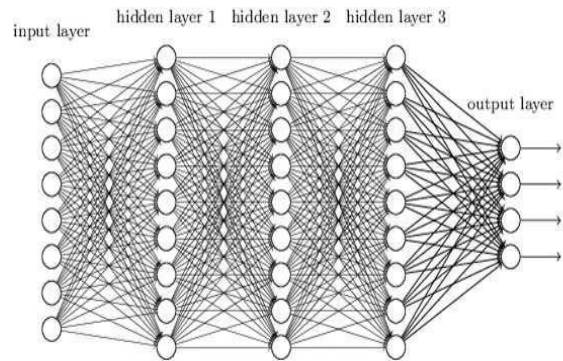

*Fig. 5: Deep Neural Network*

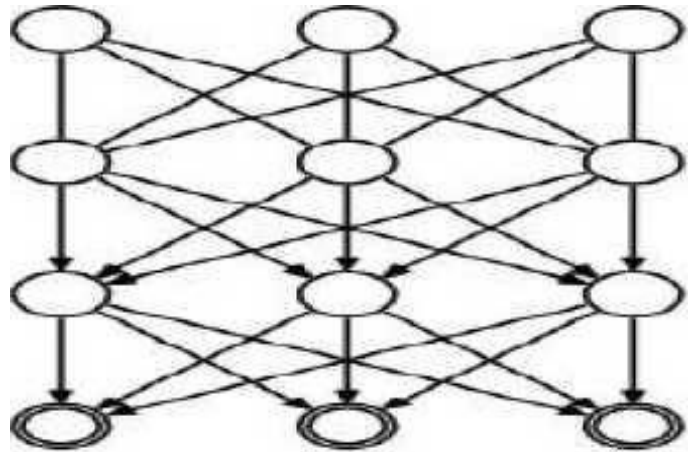

**Fig. 6: Deep Belief Network (BDN)**

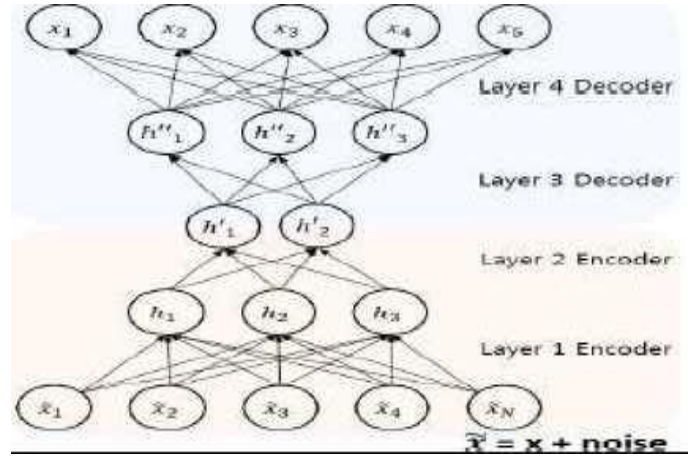

**Fig. 7: Deep Autoencoder (dA)**

## 3.1 Dataset

Deep learning requires massive amount of training dataset as classification accuracy of deep learning classifier is largely dependent on the quality and size of the dataset, however, unavailability of dataset is one the biggest barrier in the success of deep learning in medical imaging. On the other hand, development of large medical imaging data is quite challenging as annotation requires extensive time from medical experts especially it requires multiple expert opinion to overcome the human error. Furthermore, annotation may not be possible due to unavailability of qualified expert or availability of suf-ficient cases are also issue in case of rare disease. Another issue major issue is unbalancing of data that is very common in health sector i.e.rare diseases, by virtue of being rare, are underrepresented in the data sets. If not accounted for properly, the class imbalance that ensues.

## 3.2 Privacy and Legal Issue

It is much more complicated and difficult to share the medical data as compared to real world images. Data privacy is effectively both sociological as well as technical issue, which must be addressed jointly from both perspectives. HIPAA comes to mind when privacy is discussed in health sector. It provides legal rights to patients regarding their personally identifiable information and establish obligations for healthcare providers to protect and restrict its use or disclosure. While the rise of healthcare data, data analytics researchers see big challenges that how to anonymize patient information to prevent its use or disclosure ? Discarding such information (social security number, medical record number, name and age) make it complex to link the data up to unique individual. But still, an attacker can identify somehow using association. The other way is differential privacy which restrict the data to organization based on requirement of data need. These privacy challenges are factors that can lead to situations where, data analytics model likely to impact it negatively from both legal as well as ethical perspective. The main privacy challenges associated with healthcare data analytics, overrunning the privacy concerns of traditional data processing, are as follows:

One important issue to address that how to share sensitive data of data while limiting disclosure and limiting its sharing by ensuring the sufficient data utility i.e. Year of birth, 3-digit Zip code, gender is unique for 0.04% of US population while Date of birth, 5-digit Zip code and gender is unique for 87% of US population The limited restriction data access, unfortunately reduce information content too that might be very important. In addition to this, we don't have static data but its size is increasing over time thus none of the existing prevailing methods results in making data secure.

## 3.3 Data Interoperability and Data Standards

Data interoperability and data standards are the one of major barrier. Currently, nature of data differ from hardware to hardware thus there exist large variation in images due to sensors and other factors. Furthermore, the breadth of any applications medical sector requires to combine several dif-ferent datasets for better algorithms learning and accuracy. Interoperability is backbone to critical improvements in health sector and yet has to become a reality. Similar to the concept of ATM network, Health data should be standardized and shared between providers. To achieve interoperability level, HIPAA, HL7, HITECH and other health standardization bodies have defined some standards and guidelines. How an organization gets to know that if it meet interoperability and security standards. Authorized testing and certi-fying body (ATCB) provides an independent, third party opinion on EHR. Two types of certification (CCHIT and ARRA) are used to evaluate the sys-tem. Review process includes standardized test scripts and exchange test of standardized data.

## 3.4 Black Box and Deep Learning

Medical imaging broke paradigms when it first began more than 100 years ago and deep learning algorithms gave new birth to medical imaging appli-cation and open new possibilities. It solves the problems previously thought to be unsolvable by a machine learning algorithms, however, deep learning is not free form the problems. One of the biggest issues is so called black-box problem, although math used to construct a neural network is straightfor-ward but how the output was arrived is exceedingly complicated i.e. machine learning algorithms get bunch of data as input, identify patterns and build predictive model but understanding how the model worked is issue. The deep learning model is offer uninterpretable and most of the researchers are using it without know the working process that why it provides better result.

## 4 Deep Learning in Medical Imaging

Many image diagnosis task requires initial search to identify abnormalities, quantify measurement and changes over time. Automated image analysis tool based on machine learning algorithms are the key enablers to improve the quality of image diagnosis and interpretation by facilitating through efficient identification of finding. Deep learning is one extensively applied techniques that provides state of the aft accuracy. It opened new doors in medical image analysis that have not been before. Applications of deep learning in healthcare covers a broad range of problems ranging from cancer screening and disease monitoring to personalized treatment suggestions. Various sources of data today - radiological imaging (X-Ray, CT and MRI scans), pathology imaging and recently, genomic sequences have brought an immense amount of data at the physicians disposal. However, we are still short of tools to convert all this data to useful information. In the below discussion, we highlighted state of the art applications of deep learning in medical image analysis. Though, the list is by no means complete however it provides an indication of the long-ranging deep learning impact in the medical imaging industry today.

## 4.1 Diabetic Retinopathy

Diabetes Mellitus (DM) is a metabolic disorder in which pancreases cannot produce proper insulin (Type-1) or the body tissues do not response to the insulin properly (Type-2) which results in high blood sugar. The Diabetic Retinopathy (DR) is an eye disease due to diabetes which results in eye blindness with the passage of time of diabetes in a person. According to [?], almost 415 million people are suffering from diabetes in the world and 15% amongst them have high risk of vision impairment, blindness and loss. This disease can be controlled and cure easily if it is detected on time and at early stage by retinal screening test.

Manual process of detection of DR is difficult and time consuming process at presence due to unavailability of equipment and expertise. As this disease shows hardly any symptoms in early stage and a clinician needs to examine the colored fundus image of retina which lead to delay the treatment, miscommunication and loss of follow up. Automated detection of DR based on deep learning models has proven their optimized and better accuracy. In this section, we are presenting the research work using deep learning approaches.

Gulshan et al. applied Deep Convolutional Neural Network (DCNN) on Eye Picture Archive Communication System (EyePACS-1) dataset and Messidor-2 dataset for classification and detection of moderate and worse referable [21]. The EyePACS-1 consists of approximately 10,000 retinal images and the Messidor-2 data set consists of 1,700 images that collected from 874 patients. The authors claimed 97.5%sensitivity and 93.4% specificity on EyePACS-1; and 96.1% sensitivity and 93.9% specificity on Messidor-1, respectively. Kathirvel [9] trained DCNN with dropout layer techniques and tested on publically available datasets like kaggle fundus, DRIVE and STARE for classification of fundus. The reported accuracy is up to 94-96%. Pratt et al. [6] employed NVIDIA CUDA DCNN library on Kaggle dataset consisting of above 80,000 digital fundus images. They also validated the network on 5,000 images. The images resized into 512x512 pixels and then sharpened. Finally, the features vector fed to Cu-DCNN. They classified the images into 5 classes using features like exudates, haemorrhages and micro-aneurysms and achieve upto 95% specificity, 30% sensitivity and 75% accuracy.

Haloi [7] implemented five layers CNN with drop out mechanism for detec-tion of early stage DR on Retinopathy Online Challenge (ROC) and Massidor datasets and claimed t Sensitivity, Specificity, accuracy and area under the curve (AUC) up to 97%, 96%, 96% and 0.988 on Maddissor dataset and AUC up to 0.98 on ROC dataset. Alban [8] de-noised the angiograph images of EyePACS and then applied CNNs for detection of DR. They diagnosed five classes severities and provide 79% AUC and 45% accuracy. Lim et al. [14] extracted features from identified regions using method proposed in [26] and then the features vector passed to DCNN for classification. They realized the model on DIARETDB1 and SiDRP datasets. All above works summarized in Table tableDiabetic.

*Table 2: Summary of Deep Learning (DL) for Diabetic Retinopathy (DR).*

| Authors | Model | Data set | accuracy:acc or sensitivity:sensi or specificity:spec (%) |
|---------|-------|----------|------------------------------------------------------------|
| Gulshan et al. | Deep Convolutional Neural Network | EyePACS-1 Messidor-2 | 97.5% sensi & 93.4% spec 96.1% sensi & 93.9% spec |
| Kathirvel | CNN with dropout layer | Kaggle-fundus, DRIVE and STARE | 94-96% |
| Pratt et al. | Cu-DCNN library | Kaggle | 75% acc |
| Haloi et al. | Five layers CNN | Massidor ROC | 98% AUC 97% AUC |
| Alban et al. | DCNN | EyePACS | 45% acc |
| Lim et al. | DCNN | DIARETDB1 SiDRP | – |

## 4.2 Histological and Microscopical Elements Detection

Histological analysis is the study of cell, group of cells and tissues. When different changes come at cellular and tissue level then microscopic changes, characteristics and features can be detected through microscopic image technology and stains (colorful chemicals) [24] [30] [29]. It involves number of steps like fixation, sectioning, staining and optical microscopic imaging. Different skin disease especially squamus cell carcinoma, and melanoma Other dis-eases like gastric carcinoma, gastric ephitilial metaplasia, breast carcinoma, malaria, Intestinal parasites and TB etc. Genus plasmodiums parasite is the main reason of Malaria. Microscopical imaging is the standard method for de-tection of parasites in stained blood smear sample. Mycobacteria in sputum is the main cause of Tuberculosis (TB). Smear microscopy and fluorescent auramine-rhodamine stain or Ziehl-Neelsen (ZN) stain are golden standanrds for detection TB.

Recently, the HistoPhenotypes dataset published [31] where DCNN classifier applied for diagnosis nuclei of cells of colon cancer using stained histological images. Bayramoglu and Heikkil [1] conducted two studies for detection of thoraco-abdominal lymph node and interstitial lung disease using transfer learning (fine tuning) approach with CNN model. Due to limited histological data in [31], features vector extracted using facial images [13] and natural images of ImageNet [11] using source CNN and then transferred to object CNN model for classification. The CNN classifier employed for grading gas-tric cancer by analyzing signet ring cells in tissues and epithelial layers of tissue. They also count the mitotic figures for breast cancer [16].

In [23], authors employ shape features like moment and morphological to pre-dict malaria, tuberculosis and hookworm from blood, sputum and stool sam-ples. Automatic microscopic image analysis performed using DCNN model as a classifier and reported AUC 100% for Malaria and 99% for tuberculosis and hookworm. DCNN also applied for diagnosis of malaria in [23] and intestinal parasites in [20]. Fully CNN Deep learning has been used in [36] for auto-matic cell counting. Qiu et al. also employed DCNN for detection of leukemia in metaphase [22]. Quin et al. conducted experiment using DCNN for detec-tion of malaria in thick blood smear,, intestinal parasite like helminthes in stool and mycobacteria in sputum [23]. Malaria detection is a crucial and important research area. In 2015, 438,000 people were died due to malaria according to World Health Organization. Dong et al. developed four systems for detection infected and non-infected cells by malaria using CNN mod-els and SVM. Three architectures of CNN named as GoogLeNet , LeNet-5 and AlexNet used for automatic features extraction and classification and reported 98.13%, 96.18% and 95.79% respectively. SVM based system got lowest accuracy upto 91.66%. Summary of all above written work is given in Table 3.

*Table 3: Summary of Deep Learning (DL) for Histological and Microscopical Elements Detection.*

| Authors | Model | Data set | accuracy:acc or sensitivity:sensi or specificity:spec (%) |
|---|---|---|---|
| Bayramoglu and Heikkil | transfer approach with CNN | ImageNet (source for features) HistoPheno-types dataset | – |
| quinn et al. | DCNN and shaped features like moment and morphological | microscopic image | 100% for Malaria; 99% for tu-berculosis and hookworm |
| Qiu et al. | DCNN | – | – |
| Dong et al. | GoogLeNet , LeNet-5, and AlexNet | | 98.66%, 96.18% and 95.79% acc |

4.3 Gastrointestinal (GI) Diseases Detection

Gastrointestinal (GI) consists of all organs involve in digestion of food and absorption of nutrients and excretion of the waste products. It starts from mouth to anus. The organs are esophagus, stomach, large intestine (colon or large bowel) and small intestine (small bowel). The GI may also divide into upper GI tract and lower GI tract. The upper GI tract consists of esophagus, stomach and duodenum (part of small bowel) and lower GI tract consists of most of small intestine (jejunam and jilium) and large intestine. The food digestion and absorption is affected due to different ailments and diseases like inflammation, bleeding, infections and cancer in the GI tract [23]. Ul-cers cause bleeding in upper GI tract. Polyps, cancer or diverticulitis cause bleeding from colon. Small intestine has diseases like Celiac, Crohn, malig-nant and benign tumor, intestinal obstruction, duodenal ulcer, Irritable bowel syndrome and bleeding due to abnormal blood vessels named as arteriove-nous malformations (angiodysplasias or angioectasias).

Image processing and machine learning play vital role in diagnosing and analyzing these diseases and help the doctors in making fast decision for treatment efficiently and accurately. Due to advancement in computer aided diagnosis (CAD) systems, various kind of imaging tests is in practice for digestive systems disease detection and classification. These imaging test are Wireless Capsule endoscopy, Endoscopy and enteroscopy, colonoscopy or sigmoidoscopy, Radioopaque dyes and X-ray studies, deep small bowel en-teroscopy, intra operative enteroscopy, Computed tomography and magnetic resonance imaging (MRI).

Jia et al. employed DCNN for detection of bleeding in GI 10,000 Wireless Capsule Endoscopy (WCE) images [7]. The WCE is a non-invasive image video method for examination small bowel disease. They claimed F measure approximately to 99Pei etal. mainly focused on evaluation of contraction fre-quency of bowel by investigation diameter patterns and length of bowel by measuring temporal information [19]. The authors implemented Fully Con-volutional Networks (FCN) and stacked FCN with LSTM using small and massive datasets. FCN-LSTM trained on small dataset consisted of 5 cine-MRI sequences without labeling and FCN system realized on massive dataset consisted of fifty raw cine MRI sequence with labeling Wimmer et al. learned features from ImageNet dataset and then the learned feature vector fed to CNN SoftMax for classification and detection of celiac disease using duode-nums endoscopic images [33].

A popular approach of automatic feature extraction from endoscopy images adopted using CNN [38]. Then the features vector to the SVM for classifica-tion and detection of gastrointestinal lesions. The proposed system realized on 180 imagesfor lesions detection and 80% accuracy reported. Similarly hy-brid approach used by [5]. Fast features extraction using CNN architecture in [29] applied and then the extracted features passed to SVM for detection of inflammatory gastrointestinal disease in WCE videos. The experiments con-ducted on 337 annotated inflammatory images and 599 non-inflammatory images of the GI tract of KID [30]. Training set containing 200 normal and 200 abnormal while test set containing 27 normal and 27 abnormal and ob-tained overall accuracy upto 90%.

The work involved in detection of polyp in colonoscopy videos using rep-resentation of image in three ways [32]. Number of CNN models trained on isolated feature like texture, shape, color and temporal information in multiple scales which enhance the accurate localization of polyp. and then combined the results for final decision. They claimed that their polyp dataset is the largest annotated dataset and they decrease the latency of detection of polyps as compare with the stat of the art techniques. Ribeiro et al. [32] also conducted three experiment using different CNN. They applied normalization (see details in [33]). The size of dataset increased using data augmentation by making different variations of images. They presented another pixels and CNN based work [34] for prognosis of polyp tumor staging using colonic mucosa as a target attribute. The work available on GI are summarized in Table 4.

## 4.4 Cardiac Imaging

Deep learning has provided extremely promising result for cardiac imaging especially for Calcium score quantification. Number of diverse applications has been developed, CT and MRI are the most used imaging modality whereas common task for image segmentation are left ventricle. Manual identification of CAC in cardiac CT requires substan- tial expert interaction, which makes it time-consuming and in- feasible for large-scale or epidemiological studies. To overcome these limitations, (semi)-automatic calcium scoring methods have been proposed for CSCT. Recent work on Cardiac images is focusing on CT angeographic images based CAC computation using deep conventional neural network as shown in figure 9

## 4.5 Tumor Detection

When cells of any part of the body have abnormal growth and make a mass then it is called Tumor or Neoplasm. There are two types of tumor. One is non-cancerous (Benign tumor) and other is cancerous (Malignant tumor). Benign tumor is not much dangerous and it is remained stick to one part of the body and do not spread to other part of the body. While malignant tu-mor is very harmful and it spread to other part of the body. When it spreads to other part then it is difficult to treat and prognoses also become very poor.

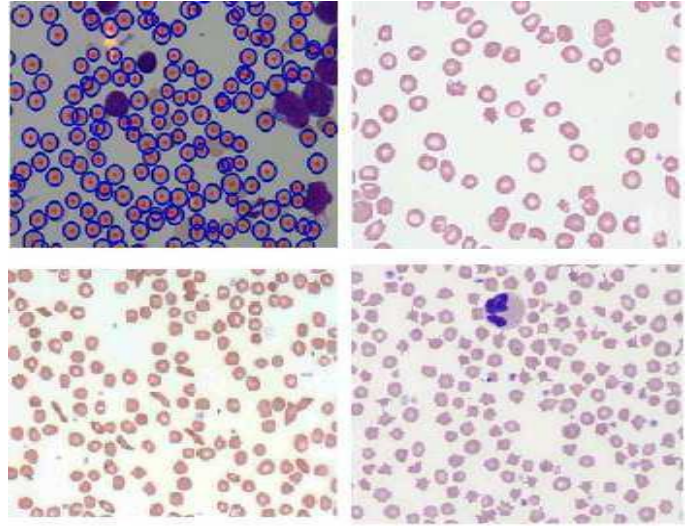

*Fig. 8: Microscopic blood smear images*

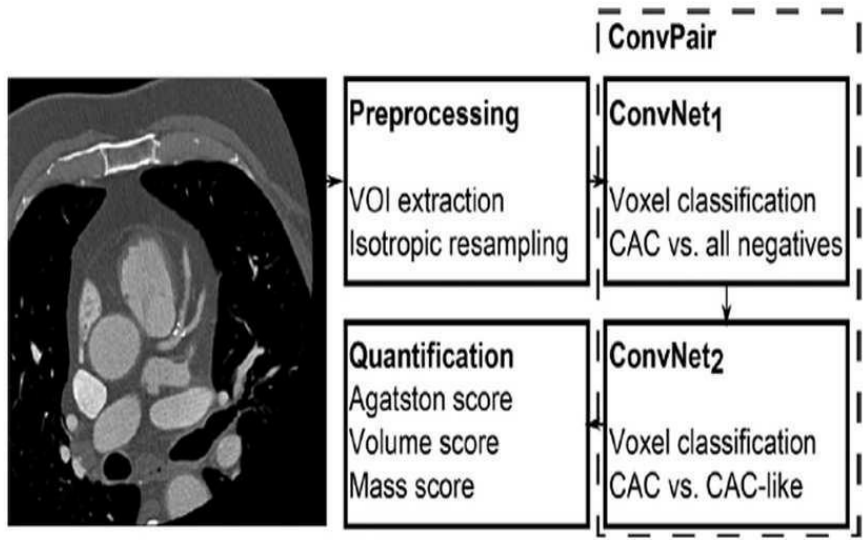

*Fig. 9: calcium score score classification from CT [34]*

Wang et al [8] used 482 mammographic images having ages from 32 years to 70 years. Images of 246 women affected by tumor. First the images de-noised using median filter and then segment the breast tumor using region growth, morphological operations and modified wavelet transformation. Then morphological and textural features passed to extreme learning machine and SVM for classification and detection of breast tumor. The total error rate was 84 using ELM and 96 using SVM. Authors in [35] have limited data of malignant mass and benign solitary cysts. CNN model needs large amount of images for better finding of cysts and mass. Therefore CNN fed by different variations of images and reported Area under curve upto 87%.

Arevalo et al.[36] conducted experiment on a benchmark dataset having 736 mediolateral oblique and craniocaudal mammographic views from 344 cancerous patients. They segmented the images manually into 310 malignant and 426 benign lesions. First, the images enhanced then fed to CNN for identification of benign and malignant lesions. They reported 82.6% AUC. Huynh et al. [37] used CNN for features learning on breast ultrasound images having 2393 regions of interests form 1125 patients. They perform two experiments. In first experiment, SVM model classified the extracted features into malignant, benign and cystic and got satisfactory result. In second experiment, SVM classified on hand crafted features. They obtained 88% AUC on CNN features and 85% AUC on hand crafted features. Antropova et al. [38] investigated CNN for transfer learning of features form ImageNet dataset (non-madical) and SVM applied on 4,096 extracted features for classification of breast lesions as malignant and benign using 551 MRI images consisting of 194 benign and 357 malignant. AUC reported upto 85%.

SVM used for classification and CNN used for features extraction in [39]. They obtained AUC 86% on dataset containing 219 lesions in 607 breast images. In [40], frozen approach of transfer learning used. DCNN trained on mammographic images with drop out and jittering approaches with 99% AUC and then validate on DTB images with 90% AUC after transfer learn-ing. The datasets containing 2282 digitized film and digital mammograms [41] and 324 DBT volumes [42]. Training set containing 2282 with 2461 le-sions and 230 DBT views with 228 masses. The remaining images used as independent test. Shin et al.[43] conducted fine tuning approach of transfer learning on ImageNet using CNN. Then CNN model applied as a classifier for detection of lesion in thoraco-abdominal lymph node and interstitial lung disease. The authors got sensitivity upto 83% and 85% and AUC upto 94% and 95%, respectively. A brief summary of published work is given in Table 5.

## 4.6 Alzheimer's and Parkinsons Diseases Detection

Parkinsons disease (PD) is a neurological disorder associated with a progressive decline in motor precision and sensor motor integration stemming presumably from a disorder of the basal ganglia [46]. Parkinsons disease is associated with the breaking up or the dying of dopaminergic neurons. Neurological testing like MMSE [47], and UPDRS [48]) and brain scans are routinely used to determine the diagnosis of AD [49]. The scale and shift invariant based features like shape of data, mean and standard deviation using CNN model (LeNet-5) , carries out classification on functional MRI 4D in work of Sarraf and Tofigh [50]. The proposed system trained on 270900 images and validated and tested on 90300 images in fMRI. The authors obtained 96.86% accuracy for detection of affected brains by Alzheimer disease.

Suk [51] employed Deep Boltzmann Machine (DBM) for features extrac-tion and detection of abnormalities from 3D patch of MRI and PET images. The results were validated on ADNI dataset [52] for alzheimers disease us-ing dataset of PET, MRI and combination of PET and MRI and obtained accuracies upto 92.38%, 92.20% and 95.35%. Hosseini-Asl [53] also explored 3D-CNN for diagnosis AD and extracted generic features using CADDemen-tia MRI dataset. The authors then fine tuned three fully connected layers CNN for classification of AD using ADNI dataset. Sarraf et al. [54] diag-nosed Alzheimer disease in adults (above 75 years old) using fMRI and MRI images. The authors conducted studies based on research and clinical appli-cations. CNN model employed for detection of healthy or Alzheimers brain and report 99.9 for functional MRI data and 98.84% for MRI data, respec-tively. They then performed classification on subject level and finally decision making based algorithm applied. The final accuracy improved upto 97.77% and 100% for fMRI and MRI subjects. In [55], sparse auto encoder (a neural network) used for extraction of features and then 3D-CNN applied as a clas-sifier on ADNI dataset consisting of neuron images. The dataset is divided into training set (1,731 samples), validation set (306 samples) and test set (228 sample) and achieved performance upto 95.39% for AD and 95.39%.

Liu et al. [? ] also used sparse auto encoder fro extraction generic features and then applied sofmax of CNN for classification of affected brain by Alzheimer; or, prodromal stage or mild stage of Alzheimer. They achieved accuracy upto 87.76% on binary images of MRI and PET for early stage detection of Alzheimer disease. All methods are summarized in Table 6.

*Table 4: Summary of Deep Learning (DL) for GI.*

| Authors | Model | Data set | accuracy:acc or sensitivity:sensi or specificity:spec (%) |
|---|---|---|---|
| Bayramoglu and Heikkil | transfer approach with CNN | ImageNet (source for features) HistoPhenotypes dataset | — |
| quinn et al. | DCNN and shaped features like moment and morphological | microscopic image | 100% for Malaria; 99% for tuberculosis and hookworm |
| Qiu et al. | DCNN | — | — |
| Dong et al. | GoogLeNet , LeNet-5, and AlexNet | | 98.66%, 96.18% and 95.79% acc |

*Table 5: Summary of Deep Learning (DL) for Tumor Detection.*

| Authors | Model | Data set | accuracy:acc or sensitivity:sensi or specificity:spec (%) |
|---|---|---|---|
| Bayramoglu and Heikkil | transfer approach with CNN | ImageNet (source for features) HistoPhenotypes dataset | — |
| quinn et al. | DCNN and shaped features like moment and morphological | microscopic image | 100% for Malaria; 99% for tuberculosis and hookworm |
| Qiu et al. | DCNN | — | — |
| Dong et al. | GoogLeNet , LeNet-5, and AlexNet | | 98.66%, 96.18% and 95.79% acc |

*Table 6: Summary of Deep Learning (DL) for Alzheimer Disease Detection.*

| Authors | Model | Data set | accuracy:acc or sensitivity:sensi or specificity:spec (%) |
|---|---|---|---|
| Bayramoglu and Heikkil | transfer approach with CNN | ImageNet (source for features) HistoPhenotypes dataset | — |
| quinn et al. | DCNN and shaped features like moment and morphological | microscopic image | 100% for Malaria; 99% for tuberculosis and hookworm |
| Qiu et al. | DCNN | — | — |
| Dong et al. | GoogLeNet , LeNet-5, and AlexNet | | 98.66%, 96.18% and 95.79% acc |

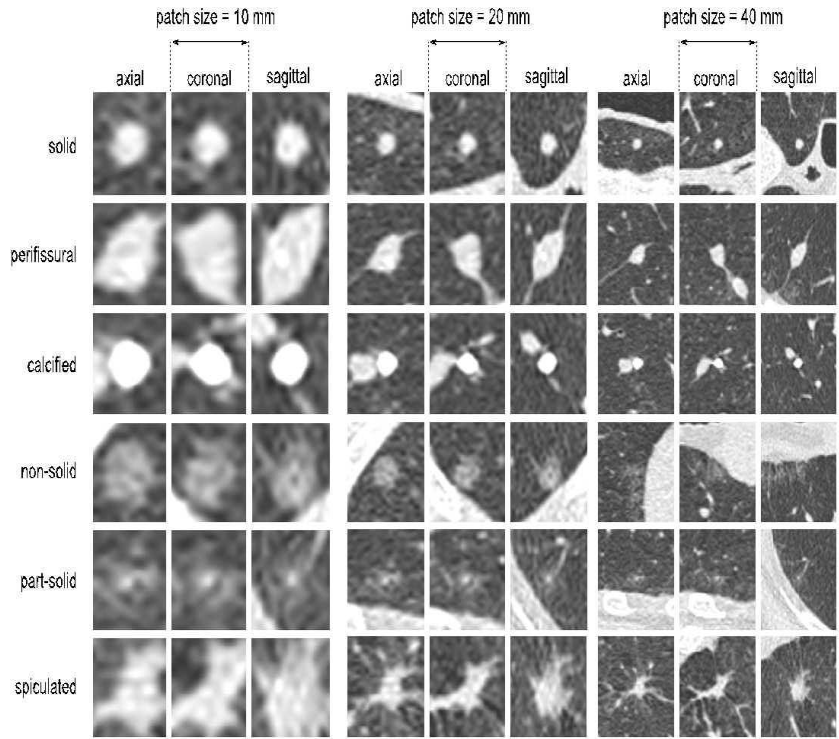

*Fig. 10: Lungs nodules segmentation [3]*

| Application Area | Input Data | Deep Learning Method |
|---|---|---|
| Cardiac CAC | CT | CNN [12, 34, 35]% |
| Lungs cancer | MRI | CNN [25] |
| Lungs cancer | CT | DNN [2, 18] |
| Diabetic retinopathy | fundus Image | CNN [6, 21] |
| Blood Analysis | Microscopic | CNN [24, 36] |
| Blood Analysis | Microscopic | DBN [4] |
| Blood vessel | Fundus | DNN [15] |
| Blood vessel | Fundus | CNN [17] |
| Brain lesion segmentation | MRI | CNN [3, 8, 10] |
| Polyp Recognition | Endoscopy | CNN [28, 33, 37] |
| Alzheimer's disease | PET | CNN [27] |

Table 7: Summary of some deep learning based medical imaging

# 5 Open Research Issues and Future Directions

Three trends that drive the deep learning revolution are availability of big data, recent deep learning algorithm modeled on the human brain and pro-cessing power. While deep learning potential benefits are extremely significant and so are the initial efforts and costs. Big companies like Google DeepMind, IBS Watson, research labs along with leading hospitals and vendors are com-ing together and working toward the optimal solution of big medical imaging. Siemen, Philips, Hitachi and GE Healthcare etc. have already made signifi-cant investments. Similarly research lab such as Google, IBM are also invest-ing towards the delivery of efficient imaging applications i.e. IBM Watson is working with more than 15 healthcare provides to learn how deep learning could work in real world.Similary google DeepMind health is collaborating with NHS, UK to apply deep learning on different healthcare applications ( for example : anonymised eye scans analysis could help to find the sing of diseases that could leads to blindness) on dataset of 1.6 million patient. GE Healthcare partnership with Bostons Children Hospital is working to create smart imaging technology for detecting pediatric brain disorders. Further-more, GE Healthcare and UC San Francisco has also announced a 3-year partnership to develop a set of algorithms to differentiate between normal result and one that requires further attention by expert.

## 5.1 Requires Extensive Inter-organization Collaboration

Despite great effort done by big stakeholder and their predictions about the growth of deep learning and medical imaging, there will be a debate on re-placing human with machine however, deep learning has potential benefits towards disease diagnosis and treatment. However, there are several issues that need to be solved to make it possible earlier. Collaboration between hospital providers, vendors and machine learning scientists is extensively re-quired to windup this extremely beneficial solution for improving the quality of health. This collaboration will resolve the issue of data unavailability to the machine learning researcher. Another major issue is, we need more so-phisticated techniques to deal extensive amount of healthcare data, especially in future, when more of the healthcare industry will be based on body senor network.

## 5.2 Need to Capitalize Big Image Data

Deep learning applications rely on extremely large dataset, however, availabil-ity is of annotated data is not easily possible as compared to other imaging area. It is very simple to annotate the real world data i.e. annotation of men and woman in crowd, annotating of object in real world image. However, annotation of medical data is expensive, tedious and time consuming as it requires extensive time for expert (especially due the sensitivity of domain, annotation required opinions of different expert on same data), furthermore annotation may not be always possible in case of rare cases. Thus sharing the data resource with in different healthcare service providers will help to overcome this issue somehow.

## 5.3 Advancement in Deep Learning Methods

Majority of deep learning methods focus on supervised deep learning how-ever annotations of medical data especially image data is not always possible i.e. in case when rare disease or unavailability of qualified expert. To over-come, the issue of big data unavailability, the supervised deep learning field is required to shift from supervised to unsupervised or semi-supervised. Thus, how efficient will be unsupervised and semi-supervised approaches in medical and how we can move from supervised to transform learning without effecting the accuracy by keeping in the healthcare systems are very sensitive. Despite current best efforts, deep learning theories have not yet provided complete solutions and many questions are still unanswered, we see unlimited in the opportunity to improve.

## 5.4 Black-Box and Its Acceptance by Health Professional

Health professional wary as many question are still unanswered and deep learning theories has not provided complete solution. Unlike health profes-sional, machine learning researchers argues interoperability is less of an issue than a reality. Human does not care about all parameters and perform com-plicated decision, it is just mater of human trust. Acceptance of deep learning in health sector need proof form the other fields, medical expert, are hoping to see its success on other critical area of real world life i.e. autonomous car, robots etc. Even though great success of deep learning based method, decent theory of deep learning algorithms is still missing. Embarrassment due to the absence this is well recognized by the machine learning community. Black-box could be another one of the main challenge, legal implications of black box functionality could be barrier as healthcare expert would not rely on it. Who could be responsible if the result went wrong. Due to the sensitivity of this area, hospital may not comfortable with black-box i.e. how it could be traced that particular result is from the ophthalmologist. Unlocking of black box is big research issue, to deal with it, deep learning scientist are working toward unlocking this proverbial black box.

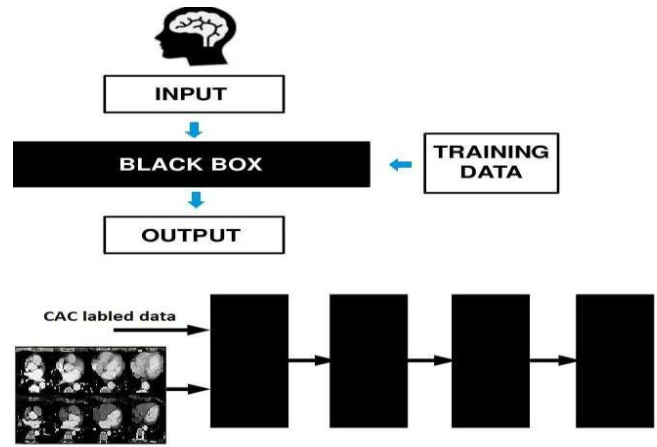

*Fig. 11: Deep Learning: A Black Box*

## 5.5 Privacy and league issues

Data privacy is effected by both sociological as well as technical issue, that need to be addressed jointly from both sociological and technical perspec-tives. HIPAA (Health Insurance Portability and Accountability Act of 1996) comes to mind when privacy is discussed in health sector. It provides legal rights to patients regarding their personally identifiable information and es-tablish obligations for healthcare providers to protect and restrict its use or disclosure. While the rise of healthcare data, researchers see big challenges that how to anonymize the patient information to prevent its use or disclosure ? The limited restriction data access, unfortunately reduce information con-tent too that might be very important. Furthermore, real data is not static but its size is increasing and changing overtime, thus prevailing methods are not sufficient.

## 6 Conclusion

During the recent few years, deep learning has gained a central position toward the automation of our daily life and delivered considerable improvements as compared to traditional machine learning algorithms. Based on the tremendous performance, most researchers believe that within next 15 years, deep learning based applications will take over human and most of the daily activities with be performed by autonomous machine. However, penetration of deep learning in healthcare especially in medical image is quite slow as compare to the other real world problems. In this chapter, we highlighted the barriers, that are reducing the growth in health sector. In last section, we highlighted state of the art applications of deep learning in medical image analysis. Though, the list is by no means complete however it provides an indication of the long-ranging deep learning impact in the medical imaging industry today. Finally, we have highlighted the open research issues.

Many big research organization are working on deep learning based solu-tion that encourage to use deep learning to apply deep learning on medical images. Looking to the brighter side of machine learning, we are hoping the sooner human will be replaced in most of the medical application especially diagnosis. However, we should not consider it as only solution as there are sev-eral challenges that reduces its growth. One of the big barrier is unavailability of annotated dataset. Thus, this question is still answerable, that whether we will be able to get enough training data without effecting the performance of deep learning algorithms. Recent development on other application showed that bigger the data, better the result, however, how big data could be used in healthcare.

So far deep learning based application provided positive feedback, however, but due to the sensitivity of healthcare data and challenges, we should look more sophisticated deep learning methods that can deal complex healthcare data efficiently. Lastly we conclude that there are unlimited opportunities to improve healthcare system.

## References

[1] Neslihan Bayramoglu and Janne Heikkil̈a. Transfer learning for cell nu-clei classification in histopathology images. In *Computer Vision– ECCV 2016 Workshops*, pages 532–539. Springer, 2016.

[2] Francesco Ciompi, Kaman Chung, Sarah J van Riel, Arnaud Arindra Adiyoso Setio, Paul K Gerke, Colin Jacobs, Ernst Th Scholten, Cornelia Schaefer-Prokop, Mathilde MW Wille, Alfonso Marchiano, et al. Towards automatic pulmonary nodule management in lung cancer screening with deep learning. *arXiv preprint arXiv:1610.09157*, 2016.

[3] Zhipeng Cui, Jie Yang, and Yu Qiao. Brain mri segmentation with patch-based cnn approach. In *Control Conference (CCC), 2016 35th Chinese*, pages 7026–7031. IEEE, 2016.

[4] Rahul Duggal, Anubha Gupta, Ritu Gupta, Manya Wadhwa, and Chirag Ahuja. Overlapping cell nuclei segmentation in microscopic images using deep belief networks. In *Proceedings of the Tenth Indian Conference on Computer Vision, Graphics and Image Processing*, page 82. ACM, 2016.

[5] Spiros V Georgakopoulos, Dimitris K Iakovidis, Michael Vasilakakis, Vassilis P Plagianakos, and Anastasios Koulaouzidis. Weakly-supervised convolutional learning for detection of inflammatory gastrointestinal le-sions. In *Imaging Systems and Techniques (IST), 2016 IEEE Interna-tional Conference on*, pages 510–514. IEEE, 2016.

[6] Varun Gulshan, Lily Peng, Marc Coram, Martin C Stumpe, Derek Wu, Arunachalam Narayanaswamy, Subhashini Venugopalan, Kasumi Widner, Tom Madams, Jorge Cuadros, et al. Development and validation of a deep learning algorithm for detection of diabetic retinopathy in retinal fundus photographs. *JAMA*, 316(22):2402–2410, 2016.

[7] Xiao Jia and Max Q-H Meng. A deep convolutional neural network for bleeding detection in wireless capsule endoscopy images. In *Engineering in Medicine and Biology Society (EMBC), 2016 IEEE 38th Annual International Conference of the*, pages 639–642. IEEE, 2016.

[8] Konstantinos Kamnitsas, Christian Ledig, Virginia FJ Newcombe, Joanna P Simpson, Andrew D Kane, David K Menon, Daniel Rueckert, and Ben Glocker. Efficient multi-scale 3d cnn with fully connected crf for accurate brain lesion segmentation. *Medical Image Analysis*, 36:61–78, 2017.

[9] C. T. R. Kathirvel. Classifying Diabetic Retinopathy using Deep Learning Architecture. *International Journal of Engineering Research Technology*, 5(6), 2016.

[10] Jens Kleesiek, Gregor Urban, Alexander Hubert, Daniel Schwarz, Klaus Maier-Hein, Martin Bendszus, and Armin Biller. Deep mri brain extrac-tion: a 3d convolutional neural network for skull stripping. *NeuroImage*, 129:460–469, 2016.

[11] Alex Krizhevsky, Ilya Sutskever, and Geoffrey E Hinton. Imagenet classi-fication with deep convolutional neural networks. In *Advances in neural information processing systems*, pages 1097–1105, 2012.

[12] Nikolas Lessmann, Ivana Isgum, Arnaud AA Setio, Bob D de Vos, Francesco Ciompi, Pim A de Jong, Matthjis Oudkerk, P Th M Willem, Max A Viergever, and Bram van Ginneken. Deep convolutional neural networks for automatic coronary calcium scoring in a screening study with low-dose chest ct. In *SPIE Medical Imaging*, pages 978511–978511. International Society for Optics and Photonics, 2016.

[13] Gil Levi and Tal Hassner. Age and gender classification using convolutional neural networks. In *Proceedings of the IEEE Conference on Com-puter Vision and Pattern Recognition Workshops*, pages 34–42, 2015.

[14] Gilbert Lim, Mong Li Lee, Wynne Hsu, and Tien Yin Wong. Transformed representations for convolutional neural networks in diabetic retinopathy screening. *Modern Artif Intell Health Anal*, 55:21–25, 2014.

[15] Pawel Liskowski and Krzysztof Krawiec. Segmenting retinal blood vessels with¡? pub newline?¿ deep neural networks. *IEEE transactions on medical imaging*, 35(11):2369–2380, 2016.

[16] Christopher Malon, Matthew Miller, Harold Christopher Burger, Eric Cosatto, and Hans Peter Graf. Identifying histological elements with convolutional neural networks. In *Proceedings of the 5th international conference on Soft computing as transdisciplinary science and technol-ogy*, pages 450–456. ACM, 2008.

[17] Lua Ngo and Jae-Ho Han. Advanced deep learning for blood vessel segmentation in retinal fundus images. In *Brain-Computer Interface (BCI), 2017 5th International Winter Conference on*, pages 91–92. IEEE, 2017.

[18] Rahul Paul, Samuel H Hawkins, Lawrence O Hall, Dmitry B Goldgof, and Robert J Gillies. Combining deep neural network and traditional image features to improve survival prediction accuracy for lung cancer patients from diagnostic ct. In *Systems, Man, and Cybernetics (SMC), 2016 IEEE International Conference on*, pages 002570–002575. IEEE, 2016.

[19] Mengqi Pei, Xing Wu, Yike Guo, and Hamido Fujita. Small bowel motil-ity assessment based on fully convolutional networks and long short-term memory. *Knowledge-Based Systems*, 121:163–172, 2017.

[20] AZ Peixinho, SB Martins, JE Vargas, AX Falcao, JF Gomes, and CTN Suzuki. Diagnosis of human intestinal parasites by deep learning. In *Computational Vision and Medical Image Processing V: Proceedings of the 5th Eccomas Thematic Conference on Computational Vision and Medical Image Processing (VipIMAGE 2015, Tenerife, Spain*, page 107, 2015.

[21] Harry Pratt, Frans Coenen, Deborah M Broadbent, Simon P Harding, and Yalin Zheng. Convolutional neural networks for diabetic retinopathy. *Procedia Computer Science*, 90:200–205, 2016.

[22] Yuchen Qiu, Xianglan Lu, Shiju Yan, Maxine Tan, Samuel Cheng, Shibo Li, Hong Liu, and Bin Zheng. Applying deep learning technology to automatically identify metaphase chromosomes using scanning microscopic images: an initial investigation. In *SPIE BiOS*, pages 97090K–97090K. International Society for Optics and Photonics, 2016.

[23] John A Quinn, Rose Nakasi, Pius KB Mugagga, Patrick Byanyima, William Lubega, and Alfred Andama. Deep convolutional neural networks for microscopy-based point of care diagnostics. *arXiv preprint arXiv:1608.02989*, 2016.

[24] Muhammad Imran Razzak and Bandar Alhaqbani. Automatic detection of malarial parasite using microscopic blood images. *Journal of Medical Imaging and Health Informatics*, 5(3):591–598, 2015.

[25] Masaharu Sakamoto and Hiroki Nakano. Cascaded neural networks with selective classifiers and its evaluation using lung x-ray ct images. *arXiv preprint arXiv:1611.07136*, 2016.

[26] Gilbert Lim Yong San, Mong Li Lee, and Wynne Hsu. Constrained-mser detection of retinal pathology. In *Pattern Recognition (ICPR), 2012 21st International Conference on*, pages 2059–2062. IEEE, 2012.

[27] Saman Sarraf, John Anderson, Ghassem Tofighi, et al. Deepad: Alzheimer s disease classification via deep convolutional neural networks using mri and fmri. *bioRxiv*, page 070441, 2016.

[28] Santi Segu´ı, Michal Drozdzal, Guillem Pascual, Petia Radeva, Carolina Malagelada, Fernando Azpiroz, and Jordi Vitri`a. Deep learning features for wireless capsule endoscopy analysis. In *Iberoamerican Congress on Pattern Recognition*, pages 326–333. Springer, 2016.

[29] Syed H Shirazi, Arif Iqbal Umar, Saeeda Naz, and Muhammad I Razzak. Efficient leukocyte segmentation and recognition in peripheral blood image. *Technology and Health Care*, 24(3):335–347, 2016.

[30] Syed Hamad Shirazi, Arif Iqbal Umar, Nuhman Ul Haq, Saeeda Naz, and Muhammad Imran Razzak. Accurate microscopic red blood cell image enhancement and segmentation. In *International Conference on Bioinformatics and Biomedical Engineering*, pages 183–192. Springer International Publishing, 2015.

[31] Korsuk Sirinukunwattana, Shan E Ahmed Raza, Yee-Wah Tsang, David RJ Snead, Ian A Cree, and Nasir M Rajpoot. Locality sensi-tive deep learning for detection and classification of nuclei in routine colon cancer histology images. *IEEE transactions on medical imaging*, 35(5):1196–1206, 2016.

[32] Nima Tajbakhsh, Suryakanth R Gurudu, and Jianming Liang. Auto-matic polyp detection in colonoscopy videos using an ensemble of con-volutional neural networks. In *Biomedical Imaging (ISBI), 2015 IEEE 12th International Symposium on*, pages 79–83. IEEE, 2015.

[33] G Wimmer, S Hegenbart, A Vecsei, and A Uhl. Convolutional neural network architectures for the automated diagnosis of celiac disease. In *International Workshop on Computer-Assisted and Robotic Endoscopy*, pages 104–113. Springer, 2016.

[34] Jelmer M Wolterink, Tim Leiner, Bob D de Vos, Robbert W van Hamersvelt, Max A Viergever, and Ivana Iˇsgum. Automatic coronary artery calcium scoring in cardiac ct angiography using paired convolu-tional neural networks. *Medical image analysis*, 34:123–136, 2016.

[35] Jelmer M Wolterink, Tim Leiner, Max A Viergever, and Ivana Iˇsgum. Automatic coronary calcium scoring in cardiac ct angiography using convolutional neural networks. In *International Conference on Medical Image Computing and Computer-Assisted Intervention*, pages 589–596. Springer, 2015.

[36] Weidi Xie, J Alison Noble, and Andrew Zisserman. Microscopy cell counting and detection with fully convolutional regression networks.

*Computer Methods in Biomechanics and Biomedical Engineering: Imag-ing & Visualization*, pages 1–10, 2016.

[37] Yixuan Yuan and Max Q-H Meng. Deep learning for polyp recognition in wireless capsule endoscopy images. *Medical Physics*, 2017.

[38] Rongsheng Zhu, Rong Zhang, and Dixiu Xue. Lesion detection of en-doscopy images based on convolutional neural network features. In *Im-age and Signal Processing (CISP), 2015 8th International Congress on*, pages 372–376. IEEE, 2015.